\documentclass[10pt, a4paper]{article}

\usepackage{lrec-coling2024} 

\usepackage{natbib}
\usepackage{multibib}
\def\@mb@citenamelist{cite,citep,citet,citealp,citealt,citepalias,citetalias}
\makeatother
\newcites{languageresource}{~}

\usepackage{graphicx}
\usepackage{soul}
\usepackage{graphicx}
\usepackage{amsfonts}
\usepackage{amsmath}
\usepackage{amssymb}
\usepackage{multirow}
\usepackage{makecell}
\usepackage{subfigure}
\usepackage{booktabs}
\usepackage{makecell}
\usepackage{graphicx}
\usepackage{booktabs}
\usepackage{multicol}
\usepackage{multirow}
\usepackage{amsmath}
\usepackage{amssymb}
\usepackage{bbding}
\usepackage{subfigure}
\usepackage{colortbl}
\usepackage{pifont}
\usepackage{listings}
\usepackage{bbm}
\usepackage{makecell}
\usepackage{bbding}
\usepackage{mathrsfs}
\usepackage{bm}

\newcommand{\ourmethod}{\textsc{m$^3$P}}
\newcommand{\ctl}{MMCL}
\newcommand{\cvlm}{CVLM}
\newcommand{\drop}{MDropNet}
\newcommand{\dataset}{\texttt{InstrMulti102}}

\usepackage{dsfont}
\usepackage{bbm}
\usepackage{amsmath}
\usepackage{booktabs}
\usepackage{amsfonts}
\usepackage{enumitem}
\usepackage{caption}
\usepackage{multicol}
\usepackage{multirow}
\usepackage{pythonhighlight}
\usepackage{float}
\usepackage{efbox}
\usepackage{subfigure}
\usepackage{hyperref}
\usepackage{cleveref}
\usepackage{colortbl}
\usepackage{url}
\usepackage{amssymb} 
\usepackage{pifont} 
\usepackage{algorithmic}
\usepackage[ruled, lined, linesnumbered, commentsnumbered, longend]{algorithm2e}
\usepackage{titlesec}

\titleformat{\section}{\normalfont\large\bfseries\center}{\thesection.}{1em}{}
\titleformat{\subsubsection}{\normalfont\normalsize\bfseries\raggedright}{\thesubsubsection.}{1em}{}
\renewcommand\thesection{\arabic{section}}
\renewcommand\thesubsection{\thesection.\arabic{subsection}}
\renewcommand\thesubsubsection{\thesubsection.\arabic{subsubsection}}
\definecolor{lightgray}{rgb}{0.9,0.9,0.9}

\usepackage{hyperref}
 \definecolor{darkblue}{rgb}{0, 0, 0.5}
  \hypersetup{colorlinks=true, citecolor=darkblue, linkcolor=darkblue, urlcolor=darkblue}

\usepackage{xstring}

\usepackage{color}

\title{\ourmethod{}: Towards Multimodal Multilingual Translation with \\ Multimodal Prompt}

\name{
\fontsize{12}{0}\selectfont Jian Yang\textsuperscript{\rm $	\spadesuit$}, Hongcheng Guo\textsuperscript{\rm $\spadesuit$}, Yuwei Yin\textsuperscript{\rm $\bigstar$}, Jiaqi Bai\textsuperscript{\rm $\spadesuit$}, Bing Wang\textsuperscript{\rm $\spadesuit$} \\ 
\fontsize{12}{0}\textbf{Jiaheng Liu\textsuperscript{\rm $\spadesuit$}}, \textbf{Xinnian Liang\textsuperscript{\rm $\spadesuit$}}, \selectfont \textbf{Linzheng Chai\textsuperscript{\rm $\spadesuit$}}, \textbf{Liqun Yang\textsuperscript{\rm $\spadesuit$\textdagger \thanks{\textdagger \, Corresponding author. \\\url{https://huggingface.co/datasets/CSJianYang/InstrMulti102}}}}, \textbf{Zhoujun Li\textsuperscript{\rm $\spadesuit$}}
}
\address{
  \textsuperscript{\rm $\spadesuit$} State Key Lab of Software Development Environment, Beihang University \\ 
  \textsuperscript{\rm $\bigstar$} Department of Computer Science, University of British Columbia \\
  \{jiaya,  hongchengguo, bjq, bingwang\}@buaa.edu.cn; yuweiyin@cs.ubc.ca \\ \{liujiaheng, xnliang, challenging, lqyang, lizj\}@buaa.edu.cn
}

\abstract{
Multilingual translation supports multiple translation directions by projecting all languages in a shared space, but the translation quality is undermined by the difference between languages in the text-only modality, especially when the number of languages is large. To bridge this gap, we introduce visual context as the universal language-independent representation to facilitate multilingual translation.
In this paper, we propose a framework to leverage the multimodal prompt to guide the \textbf{M}ultimodal \textbf{M}ultilingual neural \textbf{M}achine \textbf{T}ranslation (\ourmethod{}), which aligns the representations of different languages with the same meaning and generates the conditional vision-language memory for translation. We construct a multilingual multimodal instruction dataset (\dataset{}) to support 102 languages 
Our method aims to minimize the representation distance of different languages by regarding the image as a central language.
Experimental results show that \ourmethod{} outperforms previous text-only baselines and multilingual multimodal methods by a large margin. Furthermore, the probing experiments validate the effectiveness of our method in enhancing translation under the low-resource and massively multilingual scenario.
\\ \newline 
}

\begin{document}

\maketitleabstract

\section{Introduction}


Multilingual neural machine translation (MNMT) models relying on text data of multiple languages support diverse translation directions in a single shared model~\cite{mnmt_challenges,microsoft_wmt2021}.
Beyond that, multimodal NMT captures the visual context from relevant images of the source sentences, bringing a further enhancement of multilingual translation~\cite{universal_mmt,on_vision_matters,on_vision_features,mmt_phrase_vision,lvp}. 
As a language-agnostic semantic representation, the image plays a bridge role in translating sentences across different languages. It is intuitively promising that images can serve as a universal router in multilingual translation.



\begin{figure}[!t]
\begin{center}
\includegraphics[width=1.0\columnwidth]{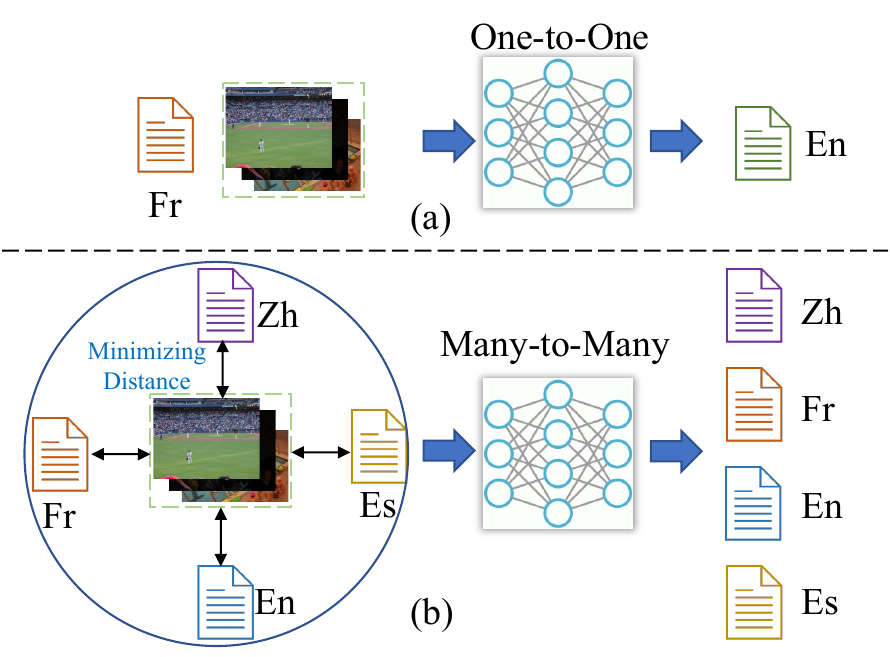}
\caption{Comparison between (a) the bilingual translation baseline and (b) our proposed \ourmethod{}.}
\label{intro}
\vspace{-10pt}
\end{center}
\end{figure}

However, previous multimodal NMT works~\cite{on_vision_matters,on_vision_features} mainly focus on the bilingual translation supervised by the image-sentence training data. In Figure~\ref{intro}(a), each bilingual model can only handle a single translation direction compared to existing thousands of languages in the world. MNMT involves more languages using available linguistic resources but only implicitly brings different languages together by sharing the same parameters. There still exists a gap between different translation directions. Some previous works~\cite{mRASP2,xmt,codeswitching_mnmt,adaptive_sparse_transformer} propose to leverage the aligned augmentation and contrastive learning across multiple languages only on the language modality. Meanwhile, images are regarded as the universal language to communicate ideas and concepts effectively across linguistic and cultural barriers~\cite{ViLBERT,cclm}. Hence, minimizing the difference across diverse directions by vision-language pair requires further exploration.

To explicitly bridge the gap among the multiple languages, we propose a multimodal prompt-based framework for multimodal multilingual neural machine translation (\ourmethod{}), which enables different translation directions between multiple source and target languages with the help of universal visual features. Specifically, we use the cross-lingual language encoder to extract the multilingual representation from the text data and the vision Transform encoder to derive the visual context. A designed multimodal prompt can be fed into the encoder-decoder model and decoder-only model (Llama2)~\cite{llama2} to verify the motivation of our work. Multimodal multilingual contrastive learning (MMCL) with masked language/image augmentation is used to align two modalities into a common semantic space. Then, we consider language representations as the query based on visual features as key and value to attend the multi-head cross-attention to generate the conditional vision-language memory (CVLM) as the encoder states. Finally, the multilingual language decoder predicts the target translation given conditional vision-language memory.


Our method is effective for multilingual translation even for the massively translation of 102 languages. Experimental results on the supervised translation directions demonstrate that our method substantially outperforms previous text-only and multilingual multimodal methods by nearly +1$\sim$+4 BLEU points. Our method is further evaluated on \dataset{} to validate the essence of the multilingual multimodal contrastive learning (MMCL). Analytic experiments emphasize the importance of alignment in both the multilingual text and vision modality, leading to better performance.

\begin{figure*}[t]
\begin{center}
\includegraphics[width=1.0\textwidth]{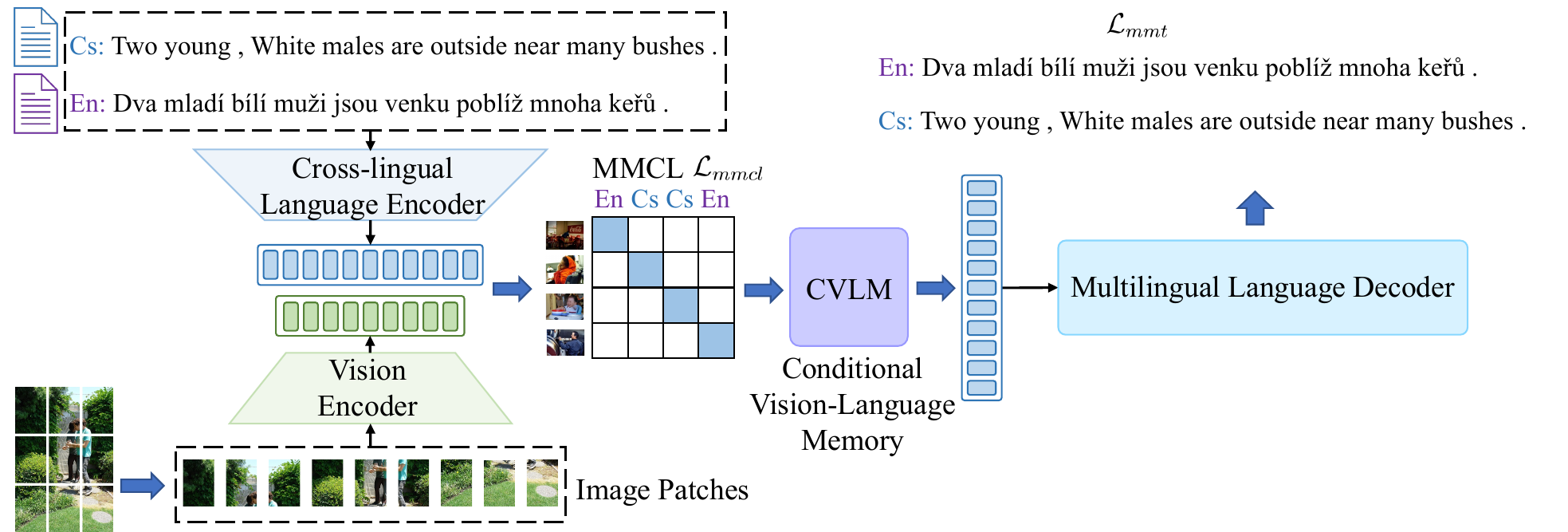}
\caption{Overview of our method. $s^{k}=\{s_u^k\}_{u=1}^{U}$ denotes the representations of the source sentence of $U$ tokens. We reshape the original image $z^k \in \mathcal{R}^{H \times W \times C}$ into $V$ patches and then encoded as $h^k=\{s_v^k\}_{v=1}^{V}$ with the vision Transformer. Given the source and visual representations $s^k$ and $h^k$, the multilingual multimodal contrastive learning (MMCL) adopted to minimize the distance between $s^k$ of different languages and $h^k$, which greatly encourages multilingual multimodal agreement in a shared space. Conditioned on the image tokens as (key,value), the language features as the query attend the multi-head attention to generate final encoder states $e^k=\{e_u^k\}_{u=1}^{U}$ as conditional vision-language for multilingual translation.}
\label{model}
\end{center}
\end{figure*}

\section{Our Method}

\subsection{Overview}
In Figure~\ref{model}, our proposed model \ourmethod{} consists of the cross-lingual language encoder, vision Transformer encoder, and the multilingual language decoder. Specifically, given the $k$-th sentence pair $(x^{k}, y^{k})$ with the image $z^k$, we first use the cross-lingual pre-trained language model to encode the source concatenation, where the target language symbol is prefixed into the source sentence to indicate the direction. Meanwhile, we reshape the image $z^k \in \mathcal{R}^{H \times W \times C}$ into a sequence of flattened patches and extract the vision context $s^k$ by the vision Transformer. To reduce the gap among different languages, the image is regarded as the central language to explicitly bring different languages to a shared semantic space using multilingual multimodal contrastive learning (\ctl{}). Then, we incorporate the language encoder states $s^{k}=\{s^k_{u}\}_{u=1}^{U}$ with $U$ tokens and auxiliary vision encoder states $h^{k}=\{h^{k}_{v}\}_{v=1}^{V}$ with $V$ tokens to generate the conditional vision-language memory (\cvlm{}) as the final encoder states. Finally, $\{e^{k}_{u}\}_{u=1}^{U}$ is fed into multilingual language decoder $\mathcal{D}$ to predict the target translation $y^k$.

\begin{figure}[ht]
\begin{center}
\includegraphics[width=1.0\columnwidth]{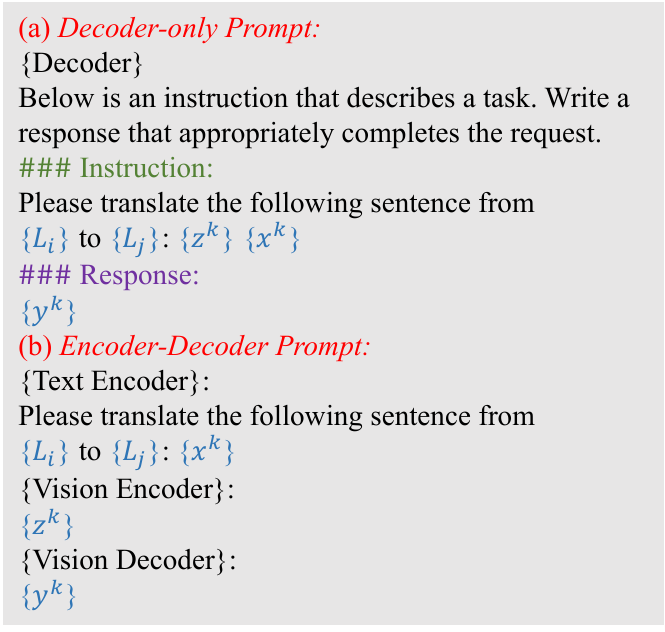}
\caption{Multimodal prompt for LLM.}
\label{prompt}
\vspace{-15pt}
\end{center}
\end{figure}

\subsection{Multilingual Multimodal Translation}

Given $M$ bilingual corpora with images $D_{all}=\{D_{m}\}_{m=1}^{M}$, where $M$ denote the number of the training corpora of $N$ languages $L_{all}=\{L_{n}\}_{n=1}^{N}$ and $L_n$ denote the $n$-th language. Each bilingual corpus with images $D_{m} = \{x^{k}, y^{k}, z^{k}\}_{k=1}^{^K}$ from $D_{all}$ consists of the source sentences, target sentences, and corresponding images. The training objective of multilingual multimodal translation can be described as:
\begin{BigEquation}
\begin{align}
\begin{split}
    \mathcal{L}_{m} &=-\sum_{m=1}^{M} \mathbb{E}_{x^{k},y^{k},z^{k}
    \in D_{m}} \left[ \log P(y^{k}|x^{k},z^{k}; \Theta) \right] 
    \label{equ:objective-mmt}
\end{split}
\end{align}
\end{BigEquation}where the multimodal multilingual model employ complete shared parameters $\Theta$ for all translation directions. We adopt Transformer as the backbone model for language and vision encoding, where the multilingual pre-trained model XLM-R~\cite{xlmr} and the pre-trained model CLIP~\cite{clip} are used to initialize the language and vision encoder. The target symbol (e.g., \texttt{[En]} or \texttt{[De]}) is prefixed to the source sentence to indicate the direction.

\paragraph{Multilingual Multimodal Prompt.}
Given the source sentence $x^{k}$, image $z^{k}$, and its translation $y^{k}$, we construct the multimodal prompt as the whole input for the decoder-only model (e.g. Llama2) in Figure~\ref{prompt}(a), where $L_{i}$ and $L_{j}$ are the source and target language. For the raw image $z^{k}$, we use the vision model to encode the image into $U$ visual tokens $h^{k}=\{h^{k}_{v}\}_{v=1}^{V}$. 
For the encoder-decoder setting in Figure~\ref{prompt}(b), we separately fed the source tokens $x^{k}$ into the text encoder and image tokens $z^{k}$ into the vision encoder. 

\subsection{Multimodal Encoding}
For the encoder-decoder setting, given the text prompt, we encode the concatenation of $U$ tokens with the language Transformer encoder to obtain the language representations $s^k$:
\begin{BigEquation}
\begin{align}
\begin{split}
    s^k=\{s^k_u\}_{u=1}^{U}=\mathcal{S}(t_{L_j}, x^k)
    \label{language_encoder}
\end{split}
\end{align}
\end{BigEquation}where $\mathcal{S}$ denotes the language encoder and the $s^k = \{s^k_u\}_{u=1}^{U}$ are language features.

Similarly, to encode the image $z^k \in \mathcal{R}^{H \times W \times C}$ with $H$ height, $W$ width, and $C$ channels, we reshape the image $z^k \in \mathcal{R}^{H \times W}$ into a sequence of flattened patches $h \in \mathcal{R}^{V \times (P^2 \times C)}$, where $P$ is the resolution of the each patch and $V = \frac{H \times W}{P^2}$ is the number of patches. Given the original image $z^{k}$, based on the Transformer encoder $ \mathcal{H}$, the source language tokens $\{s_{f}\}_{f=1}^{F}$ are extracted as:
\begin{BigEquation}
\begin{align}
\begin{split}
    h^k=\{h^k_v\}_{v=1}^{V}=\mathcal{H}(z^k)
    \label{vision_encoder}
\end{split}
\end{align}
\end{BigEquation}where $\mathcal{H}$ denotes the vision Transformer encoder and the $h^k=\{h^k_v\}_{v=1}^{V}$ are vision representations.

For the decoder-only setting, we first leverage the vision extractor to obtain the visual tokens and fill them into the prompt in Figure~\ref{prompt}. All tokens are concatenated as a whole into a large language model for the final representations. Then, we can similarly get the text representations $s^{k}$ and $h^{k}$ for the following operations.

\subsection{Multilingual Multimodal Alignment}

To effectively fuse the multilingual text and vision features, the image can be regarded as the universal language to bridge the gap among different languages. We introduce multilingual multimodal contrastive learning (MMCL) to further improve text-image alignment and multilingual text-text alignment. We use the InfoNCE objective~\cite{info_nce} to learn the correspondence between image and text. In particular, we minimize the sum of two multi-modal contrastive losses:
\begin{BigEquation}
\begin{align}
\begin{split}
\mathcal{L}_{c}=\sum_{x^k, z^k\in D_{all}}\big (f(x^k, z^k)
+ f(z^k, x^k)\big )
\label{info_nce}
\end{split}
\end{align}
\end{BigEquation}where $D_{all}$ is the multilingual dataset that contains sampled multilingual image-text pairs. $f(z^k, x^k)$ and $f(x^k, z^k)$ are the contrastive loss on image-to-text similarity and text-to-image similarity. Specifically, the image-to-text contrastive loss is:
\begin{BigEquation}
\begin{align}
\begin{split}
\label{image_to_text_nce}
 f(x^k, z^k) = -\log {\frac{ \exp\left({z^k \cdot x^{k} / \tau}\right)}{ {\sum_{x \in \{x^{k}, x^-\}}} {\exp\left({z^k \cdot z / \tau}\right)}   }}
\end{split}
\end{align}
\end{BigEquation}where $\tau$ is a temperature hyper-parameter, $x^{k}$ are \textit{positive} embedded text clips overlapping with image clip embedding $z^k$, and $x^{-}$ are \textit{negative} embedded text clips that are implicitly formed by other text clips in the training batch $B$. Symmetrically, the text-to-image loss $f(z_t, z_v)$ is defined as: 
\begin{BigEquation}
\begin{align}
\begin{split}
\label{text_to_image_nce}	
 f(z^k, x^k) = -\log {\frac{ \exp\left({z^k \cdot x^{k} / \tau}\right)}{ {\sum_{x \in \{x^{k}, x^-\}}} {\exp\left({z^k \cdot z / \tau}\right)}   }}
\end{split}
\end{align}
\end{BigEquation}

To construct the multilingual text in the training batch $B$ and balance multiple bilingual corpora, we adopt a temperature-based sampling method to collect sentences of different languages in a single batch using sampling probabilities ${q_1,\dots,q_M}$:
\begin{BigEquation}
\begin{align}
\begin{split}
\label{temperature_based_sampling}	
q_m = \frac{(|D_{m}|/|D_{all}|)^{\frac{1}{\tau}}}{\sum_{i=1}^{M}(|D_i|/|D_{all}|)^{\frac{1}{\tau}}}
\end{split}
\end{align}
\end{BigEquation}where $|D_m|$ is the size of training dataset $D_m$. The temperature gradually increases to the peak value for several epochs. The temperature is calculated by $\tau_{i}=\min(\tau, \tau_{0}+\frac{i}{\mathcal{W}}(\tau-\tau_{0}))$, where $\tau_0$ and $\tau$ separately denote the initial and peak temperature, and $\mathcal{W}$ is the number of warming-up epochs.

\subsection{Multilingual Multimodal Augmentation}
Our goal is to learn to model multilingual image-text alignment by using difficult examples in the multilingual multimodal contrastive objective. We construct negatives in our training batch by using masked language/image modeling, which are semantically similar to the original sentence.

For image augmentation, we leverage the function $\mathcal{I}(\cdot)$ to augment the original image by cropping, resizing, rotation, cutout, color distortion, Gaussian blur, and Sobel filtering. Then, we divide an image into regular non-overlapping patches and mask the chosen patches sampling from a uniform distribution as masked image modeling.

For the multilingual text, we randomly mask some random spans of contiguous tokens. For each sentence, we adopt the multilingual data augmentation $\mathcal{T}(\cdot)$ to augment the original sentence of different languages. The augmented source sentence and the image \{$\mathcal{I}(x^k)$, $\mathcal{T}(z^k)$\} with multilingual multimodal augmentation (MMA) is used to enhance the contrastive learning to learn the specific representational invariances.

\subsection{Conditional Vision-Language Memory}
Given the source concatenation $s^k=\{s^k\}_{u=1}^{U}$ from the language encoder, the language is regarded as the main input (key) with the auxiliary by the multi-head cross-attention as:

\begin{BigEquation}
\begin{align}
\begin{split}
e^k = \overset{A}{\underset{a=1}{\big\|}}  \sigma \left (\frac{(W_Q^ah^k) (W_Q^as^k)^\top}{\sqrt{C}}\right) (W_V^as^k)
\label{CVLM}
\end{split}
\end{align}
\end{BigEquation}where $\|_{a=1}^A$ is the concatenation operator of $A$ attention heads and $\sigma$ denotes the softmax operation. $W_K^a, W_Q^a$, and $W_V^a$ are respectively the corresponding linear projection matrix of the query, key, and value for $a$-th head. $C$ denotes the number of feature channels. $e^k=\{e^k\}_{u=1}^{U}$ are encoder representations, which will be fed into the decoder.

\subsection{Multilingual Generation}
Our method can be split into the text translation and the image caption task. To effectively train the text encoder, our model predicts the target words only based on the source language as below:
\begin{BigEquation}
\begin{align}
\begin{split}
    y^k_{t}=\mathcal{D}(y^k_{1:t-1},s^k;\theta) 
    \label{equ:text_translation}
\end{split}
\end{align}
\end{BigEquation}where $\mathcal{D}$ denotes the standard Transformer decoder and the $y^k_{t}$ is the $t$-th target word conditioned on the previous $t-1$ tokens $y^{k}_{1:t-1}$.

\begin{table*}[t]
\centering
\resizebox{0.8\textwidth}{!}{
\begin{tabular}{l|l|ccccccc}
\toprule
\toprule
\multicolumn{2}{c|}{}  & En$\to$Fr & En$\to$Cs & En$\to$De & Fr$\to$En & Cs$\to$En & De$\to$En & Avg$_6$ \\
\midrule
\multicolumn{9}{c}{\textit{Only Trained on Text Data}} \\
\midrule
1$\to$1 & BiNMT~\cite{transformer} & 63.3  & 33.4  & 39.9  & 54.0  & 41.1  & 43.8 & 45.9 \\ \midrule
N$\to$N & MNMT~\cite{m2m}  & 63.8   & 34.0   & 40.2 &  52.0 & 41.3 & 42.5 & 45.6 \\ \midrule
\multicolumn{9}{c}{\textit{Trained on Text and Vision Data}} \\ \midrule
1$\to$1 & BiNMT~\cite{transformer}  & 63.5   & 33.0   & 40.3 & 55.1 & 41.8 & 44.1 & 46.3 \\ 
\midrule
\multirow{6}{*}{N$\to$N} & MNMT (Gated Fusion)~\cite{on_vision_matters}  & 63.8 & 34.4  & 41.0 & 51.5 & 41.1&  43.3 & 45.8 \\
& MNMT (Concatenation)~\cite{on_vision_matters}  & 63.0   & 33.8   & 38.8    & 53.3 & 43.6 & 44.0 & 46.1 \\
& mRASP2~\cite{mRASP2}  & 63.8   & 34.4   & 41.3    & 53.2 & 44.0 & 44.5 & 46.9  \\
& Selective Attn~\cite{on_vision_features} &   63.5   & 34.4   & 41.3    & 53.2 & 44.0 & 44.5 & 46.8  \\
& LVP-M$^{3}$~\cite{lvp}             &   63.4   & 34.1   & 41.4    & 53.2 & 44.0 & 44.5 & 46.8  \\
&  \ourmethod{} (Encoder-Decoder)  &  \bf 64.8  & \bf 35.2  & \bf 41.8 & \bf 53.8 & \bf44.8  & \bf 45.0 & \bf 47.6 \\
& \ourmethod{} (Decoder-only)  &  \bf 66.4  & \bf 38.1  & \bf 43.5 & \bf 56.7 & \bf46.9  & \bf 48.1 & \bf 49.9 \\
\bottomrule
\bottomrule
\end{tabular}}
\caption{X$\to$En and En$\to$X evaluation results for bilingual (1$\to$1) and many-to-many ($N \to N$) models on the Flickr2016 test set.}
\label{tab:flickr2016_supervised}
\end{table*}


\begin{table*}[t]
\centering
\resizebox{0.9\textwidth}{!}{
\begin{tabular}{l|l|ccccc|ccccc}
\toprule
\toprule
\multicolumn{2}{c|}{}  & En$\to$Fr & En$\to$De & De$\to$En & Fr$\to$En & Avg$_4$ & En$\to$Fr & En$\to$De & Fr$\to$En & De$\to$En & Avg$_4$ \\
\midrule
\multicolumn{2}{c|}{}  & \multicolumn{5}{c|}{Flick2017} & \multicolumn{5}{c}{MSCOCO} \\
\midrule
\multicolumn{12}{c}{\textit{Only Trained on Text Data}} \\
\midrule
1$\to$1 & BiNMT~\cite{transformer}        & 55.4   & 34.1   & 39.2   & 43.4   & 43.0 & 45.8 & 32.1 & 40.6  & 34.3 & 38.2 \\ \midrule
N$\to$N & MNMT~\cite{m2m}                 & 56.8   & 34.9   & 40.3   & 44.6   & 44.2 & 45.9 & 31.9 & 41.6  & 34.6 & 38.5 \\ \midrule
\multicolumn{12}{c}{\textit{Trained on Text and Vision Data}} \\ \midrule
1$\to$1 & BiNMT~\cite{transformer}                                     & 55.8 & 34.6 & 39.6 & 43.6 & 43.4 & 45.8 & 32.3  & 41.6 & 34.4  & 38.5  \\ 
\midrule
\multirow{6}{*}{N$\to$N} & MNMT (Gated Fusion)~\cite{on_vision_matters}& 56.8 & 34.3 & 40.3 & 44.2 & 43.9 & 46.8  & 32.5  & 42.2 & 34.5 & 39.0 \\
& MNMT (Concatenation)~\cite{on_vision_matters}                        & 56.4 & 34.0 & 39.4 & 43.8 & 43.4 & 46.4  & 32.6  & 42.4 & 34.1 & 38.9 \\
& mRASP2~\cite{mRASP2}                                               & 57.0 & 35.1 & 39.6 & 44.1 & 43.9 & 47.1  & 32.7  & 42.3 & 34.8 & 39.2 \\
& Selective Attn~\cite{on_vision_features}                             & 56.6 & 34.2 & 40.3 & 44.4 & 43.9 & 46.8  & 32.5  & 42.5 & 34.3 & 39.0 \\
& LVP-M$^{3}$~\cite{lvp}                                               & 57.4 & 34.4 & 40.4 & 44.7 & 44.2 & 46.8  & 32.5  & 42.6 & 34.5 & 39.1 \\
& \ourmethod{} (Encoder-Decoder)                                        
& \bf 57.4 & \bf 35.3 & \bf 41.0 & \bf 45.6 & \bf 44.8 & \bf 46.8 & \bf 33.1  & \bf 43.2 & \bf 35.2 & \bf 39.6 \\
& \ourmethod{} (Decoder-only)                                       
& \bf 58.3 & \bf 37.2 & \bf 42.2 & \bf 46.5 & \bf 46.1 & \bf 47.4 & \bf 34.2  & \bf 44.5 & \bf 36.2 & \bf 40.6 \\
\bottomrule
\bottomrule
\end{tabular}}
\caption{X$\to$En and En$\to$X evaluation results for bilingual (1$\to$1) and many-to-many ($N \to N$) models on the Flickr2017 test set and MSCOCO test set.}
\label{tab:flickr2017_supervised}
\end{table*}


Similarly, to align the image and target language, we adopt the training objective of image caption only based on the image context as below:
\begin{BigEquation}
\begin{align}
\begin{split}
    y^k_{t}=\mathcal{D}(y^k_{1:t-1},h^k;\theta) 
    \label{equ:image_translation}
\end{split}
\end{align}
\end{BigEquation}

Given the conditional vision-language memory $e^k=\{e^k\}_{u=1}^{U}$ containing language and vision information, we adopt the standard Transformer decoder to predict the target words sequentially as:
\begin{BigEquation}
\begin{align}
\begin{split}
    y^k_{t}=\mathcal{D}(y^k_{1:t-1},e^k;\theta)
    \label{equ:our}
\end{split}
\end{align}
\end{BigEquation}where $\mathcal{D}$ is the language Transformer decoder and the $y^k_{t}$ is the $t$-th target word conditioned on the previous $t-1$ tokens $y^{k}_{1:t-1}$. In our work, our model is separately trained on the objective Eq~\ref{equ:text_translation} and Eq~\ref{equ:image_translation} for $25\%$ time and on Eq~\ref{equ:our} for $50\%$, which is denoted as the Multimodal DropNet (\drop{}).

\subsection{Training Objective}
During training, \ourmethod{} is optimized by jointly minimizing the multilingual multimodal contrastive training objective from Equation~\ref{equ:objective-mmt} and translation objective from Equation~\ref{equ:text_translation}$\sim$\ref{equ:our}:
\begin{BigEquation}
\begin{align}
\begin{split}
\mathcal{L}_{all} = \mathcal{L}_{m} + \lambda\mathcal{L}_{c}
\label{cstm}
\end{split}
\end{align}
\end{BigEquation}where $\lambda$ is the coefficient to balance the translation objective and multilingual contrastive objective.

\section{Experiments}

\subsection{Datasets}

\paragraph{Multi30k.}
We conducted experiments on the widely used Multi30k benchmark~\cite{multi30k}. The training and valid sets contain 29K and 1K sentences, respectively.
The dataset contains four languages and each sentence pair has a corresponding image, including English (En), German (De), French (fr), and Czech (Cs).
We reported the results on the Flickr2016, Flickr2017, Flickr2018, and MSCOCO test sets~\cite{mscoco,multi30k_cs}, where MSCOCO is the out-of-domain dataset with ambiguous verbs.

\subsection{Baselines}

\paragraph{Text-only Methods.}
\textbf{BiNMT}~\cite{xlmr} adopt the Transformer backbone initialized by XLM-R and then only trained on single translation direction. \textbf{MNMT}~\cite{m2m} is jointly trained on all multilingual data, where the target language symbol is prefixed to the input sentence.

\paragraph{Multimodal Methods.}
\textbf{BiNMT}~\cite{transformer} is the bilingual Transformer model concatenating the language and visual feature. \textbf{MNMT (Gated Fusion)} and \textbf{MNMT (Concatenation)}~\cite{on_vision_matters} use the visual context using the gated fusion and concatenation unit, respectively. We apply the \textbf{mRASP2}~\cite{mRASP2} on the multimodal translation with the text-only contrastive learning. \textbf{Selective Attn}~\cite{on_vision_features} use a single-head attention network to correlate words with image patches. \textbf{LVP-M$^{3}$} uses the language-aware visual prompt to guide the multimodal translation.
For a fair comparison, all the language encoders are initialized by XLM-R~\cite{xlmr} and the vision encoders are initialized by CLIP~\cite{clip}.

\begin{table}[t]
\centering
\resizebox{1.0\columnwidth}{!}{
\begin{tabular}{l|ccc|c}
\toprule
\bf Model    &  Zh$\to$En &  Hi$\to$En  & Th$\to$En & Avg$_{\bm{101}}$   \\
\midrule
Text-only MNMT               &  14.3          &   13.5    &  11.1      &  14.3 \\
MNMT (Gated Fusion)          &  15.2          &   14.3    &  12.1      &  15.4           \\
MNMT (Concatenation)         &  15.1          &   14.6    &  13.1      &  15.8           \\
\bf \ourmethod{} (Encoder-Decoder)       &  \bf 16.8 & \bf 15.2    & \bf 14.8  &  \bf 18.2   \\
\bf \ourmethod{} (Decoder-only)      &  \bf 18.2 & \bf 16.4    & \bf 16.5  &  \bf 21.2   \\
\bottomrule
\end{tabular}
}
\caption{Massively multilingual translation average results (101 translation directions) on \dataset{}.}
\label{tab:massively_mt}
\end{table}

\begin{table}[t]
\small
\begin{center}
\resizebox{1.0\columnwidth}{!}{
\begin{tabular}{l|cccc}
\toprule
\bf Model & En$\to$Fr &  En$\to$De &  Fr$\to$En & De$\to$En\\ 
\midrule
Text-only MNMT   & 63.8 & 40.2  & 52.0  & 42.5 \\
\midrule
ResNet50    &  64.2 & 40.6  & 52.3  &  43.1 \\
ResNet101   &  64.4 & 40.8  & 52.4  &  43.4 \\
ViT-B/32    &  64.8 & 41.6  & \bf 53.8  &  45.0 \\
ViT-B/16    &  65.1 & 41.8  & 53.6  &  44.8    \\
ViT-B/14    &  \bf 65.2 &  \bf 41.9 & 53.4  & \bf 45.2 \\
\bottomrule
\end{tabular}}
\end{center}
\caption{Comparison of different vision backbones (e.g., CNN and Transformer backbones) on the Flickr2016 test set.}
\label{tab:backbone_ablation}
\end{table}

\subsection{Training and Evaluation}
For the encoder-decoder setting, our model comprises a language encoder initialized by the cross-lingual language pre-trained encoder XLM-R~\cite{xlmr} and a vision encoder initialized by CLIP~\cite{clip},  We train multilingual models with Adam~\cite{adam} ($\beta_{1}=0.9$, $\beta_{2}=0.98$). For the decoder-only setting, we use the Llama2~\cite{llama2} for text generation and CLIP for vision extractor.
The learning rate is set as 5e-4 with a warm-up step of 4,000. The models are trained with the label smoothing cross-entropy with a smoothing ratio of 0.1. Our model comprises a vision encoder, language encoder, and language decoder, which all consist of 12 layers with 768 hidden size and share the same embedding matrix.
For the multilingual training, the batch size is 2048 tokens on 8 Tesla V100 GPUs. The evaluation metric is the case-sensitive detokenized sacreBLEU\footnote{\url{https://github.com/mjpost/sacrebleu}}.

\subsection{Results}

\paragraph{Flickr Test Set.}
In Table~\ref{tab:flickr2016_supervised} and~\ref{tab:flickr2017_supervised}, \ourmethod{} clearly improves multilingual baselines by a large margin in 6 translation directions. Previously, text-only MNMT underperforms bilingual translation on average. Further, \textbf{MNMT (Gated Fusion)} and \textbf{Concatenation} introduce the image as the auxiliary context to enhance translation, but these methods ignore the alignment of different languages. \textbf{mRASP2} further adopt the text-text contrastive learning scheme to close the gap among representations of different languages. \ourmethod{} extract the visual and language features with the Transformer encoder and fuse them for translation in a shared space by the \ctl{} and \cvlm{}.

\paragraph{MSCOCO Test Set.}
In Table~\ref{tab:flickr2017_supervised}, we report the performance of the previous baselines and our method on the MSCOCO test set, which is more challenging for MMT models due to the out-of-domain instances with ambiguous verbs. Therefore, it more relies on the image context for disambiguation. Our method outperforms the bilingual baseline by a large margin due to the fusion of text and image.

\begin{table}[t]
\centering
\resizebox{1.0\columnwidth}{!}{
\begin{tabular}{c|l|cc}
\toprule
ID & Flickr2016          &  En$\to$De   &  De$\to$En \\
\midrule
{\large{\ding{172}}} & \ourmethod{} (our method)                   & \bf 41.6  & \bf 45.0        \\
{\large{\ding{173}}} & {\large{\ding{172}}} - \ctl{}                     & 41.2     & 44.6       \\
{\large{\ding{174}}} & {\large{\ding{173}}} - \cvlm{}                    & 40.8     & 44.0       \\
{\large{\ding{175}}} & {\large{\ding{174}}} - \drop{}                    & 40.5     & 43.8   \\
{\large{\ding{176}}} & {\large{\ding{175}}} - Multilingual Training      & 40.1     & 43.2       \\
\bottomrule
\end{tabular}
}
\caption{Ablation study of the different modules on Flickr2016. \ourmethod{} is the final model of our method.}
\label{module_ablation}
\end{table}

\begin{figure}[t]
\centering
    \subfigure[]{
    \includegraphics[width=0.4\columnwidth]{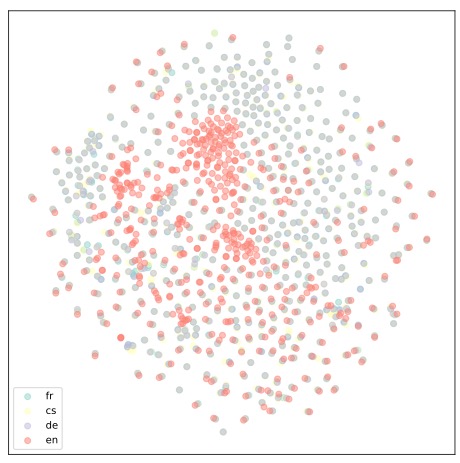}\quad
    \label{tsne_baseline}
    }
    \subfigure[]{
    \includegraphics[width=0.4\columnwidth]{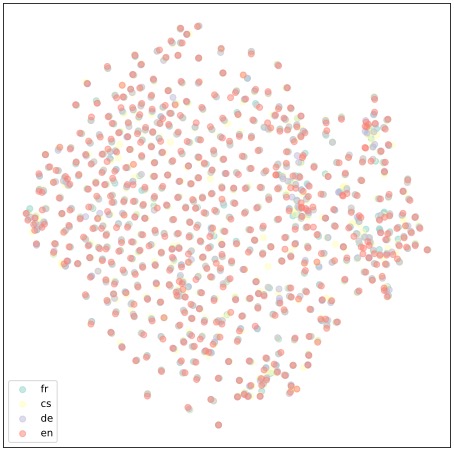}\quad
    \label{tsne_our}
    }
    \caption{Visualization of the sentence average encoder representations of all languages from the multilingual baseline (a) and our multilingual model supervised by the image context (b). Each color denotes one language.} 
    \label{tsne}
\end{figure}

\begin{figure}[t]
    \centering
    \subfigure[En$\to$Fr]{
    \includegraphics[width=0.45\columnwidth]{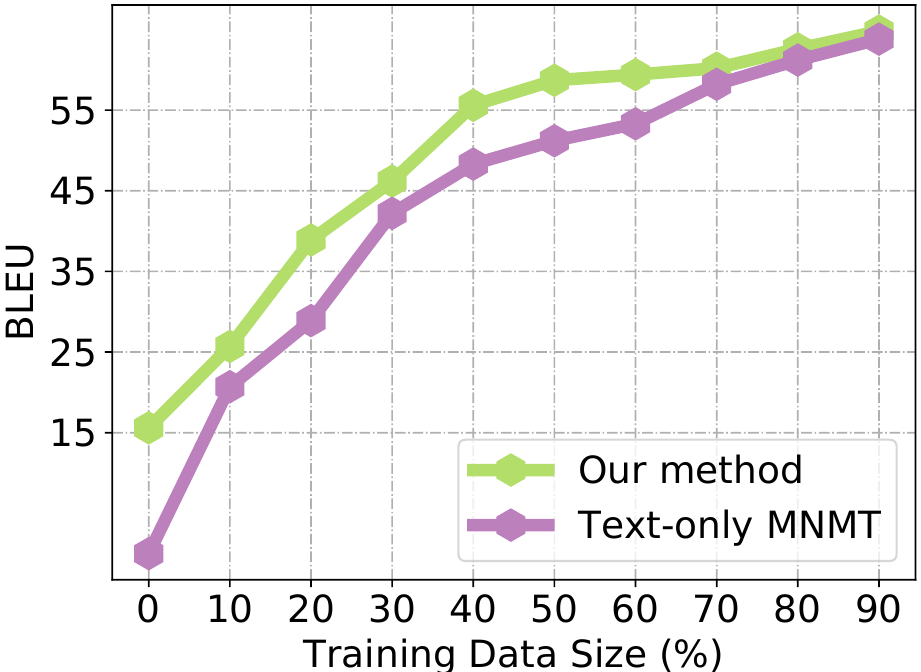}
    \label{low_resource_en2fr}
    }
    \subfigure[En$\to$De]{
    \includegraphics[width=0.45\columnwidth]{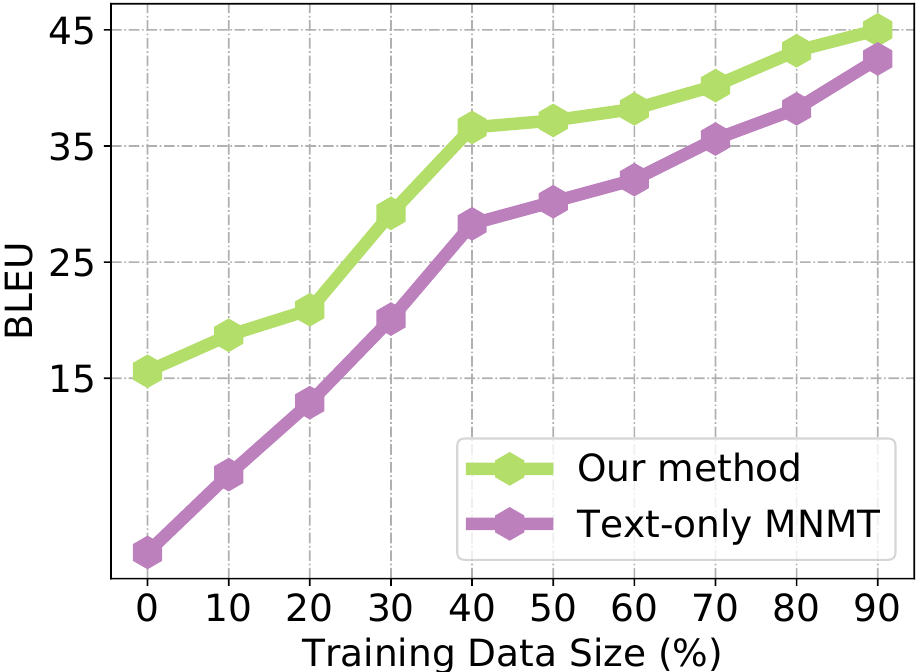}
    \label{low_resource_en2de}
    }
    \subfigure[Fr$\to$En]{
    \includegraphics[width=0.45\columnwidth]{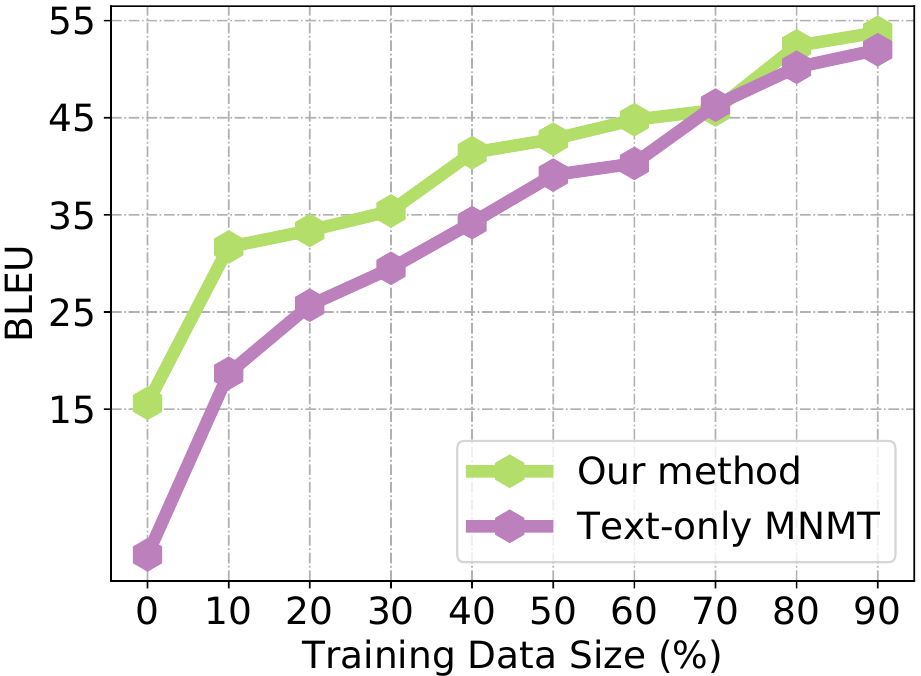}
    \label{low_resource_fr2en}
    }
    \subfigure[De$\to$En]{
    \includegraphics[width=0.45\columnwidth]{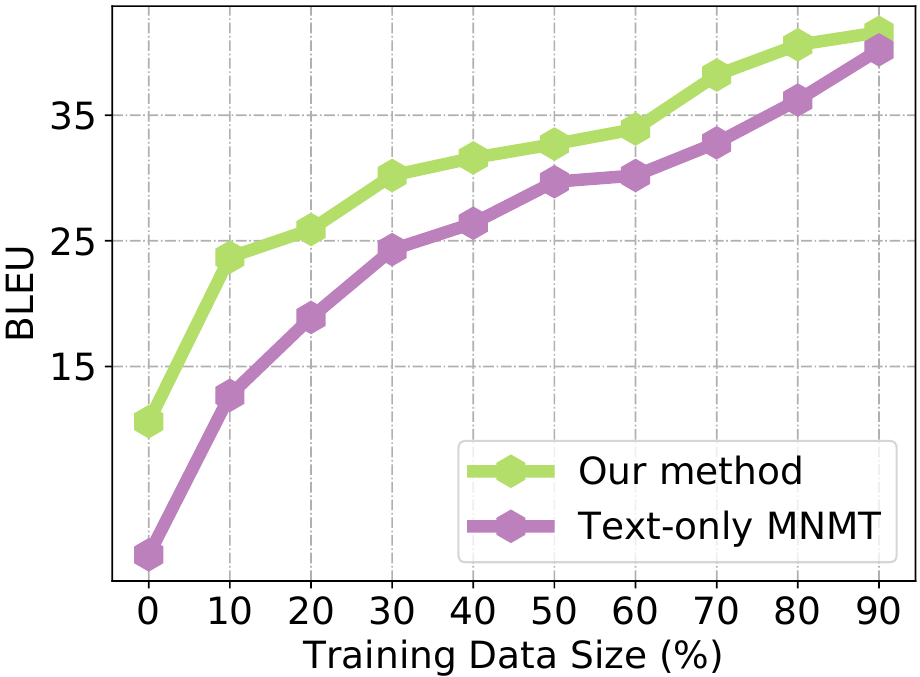}
    \label{low_resource_de2en}
    }
    \caption{The performance of our method on Flickr2016 (a) En$\to$fr, (b) En$\to$De, (c) Fr$\to$En, and (d) De$\to$En with different sizes of training data on Flickr2016.} 
    \label{low_resource}
\end{figure}

\begin{figure*}[t]
    \centering
    \subfigure[Original]{
    \includegraphics[width=0.35\columnwidth]{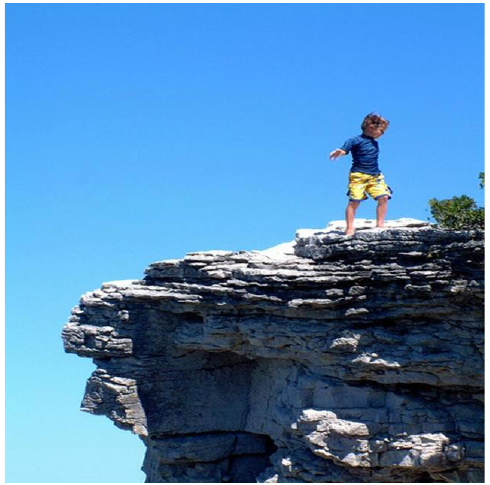}
    \label{cvlm_original}
    }
    \subfigure[En]{
    \includegraphics[width=0.35\columnwidth]{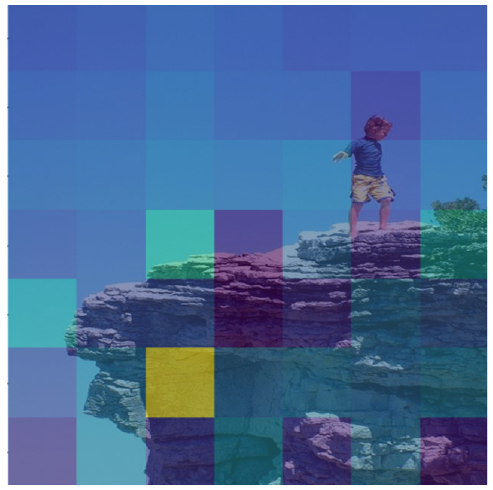}
    \label{cvlm_en}
    }
    \subfigure[De]{
    \includegraphics[width=0.35\columnwidth]{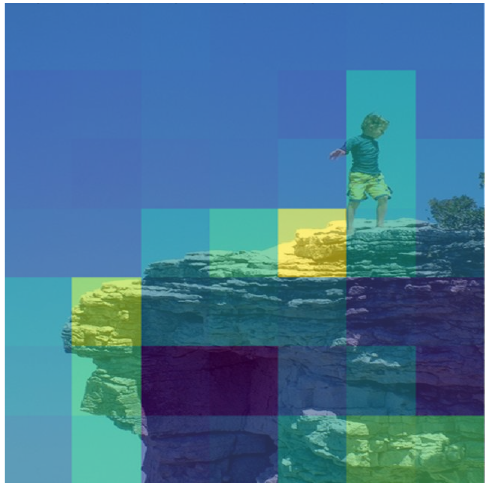}
    \label{cvlm_de}
    }
    \subfigure[Fr]{
    \includegraphics[width=0.35\columnwidth]{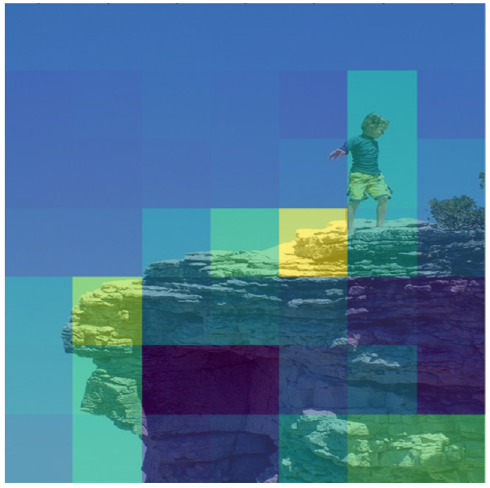}
    \label{cvlm_fr}
    }
    \subfigure[Cs]{
    \includegraphics[width=0.35\columnwidth]{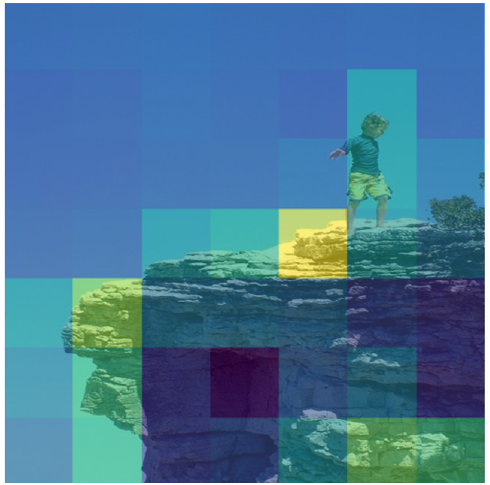}
    \label{cvlm_cs}
    }
    \caption{Representative examples of vision-language alignment from the CVLM of four languages between image patches. Brighter colors represent a higher attention value.} 
    \vspace{-10pt}
    \label{cvlm}
\end{figure*}

\section{Massively Multilingual Translation}
Considering the existing multimodal translations are limited to only a few languages, we break the limits of multilingual multimodal machine translation by extending the number of used languages in the previous benchmark Multi30k.

\paragraph{Data Construction.}
We introduce a massive multilingual multimodal machine translation dataset, called \dataset{}, originating from the previous dataset Multi30k~\cite{multi30k}. Here, we describe the details of the \dataset{}. We use the text-only multilingual Microsoft translator~\cite{microsoft_wmt2021} to construct \dataset{} by translating the English data to other 101 languages (Please refer to Appendix A for more details). The many-to-one multilingual model are jointly trained on the expanded dataset of 102 languages and then evaluated on the test set.

\paragraph{Main Results.}
In Table~\ref{tab:massively_mt}, we can see that all multilingual models with visual context perform better than the text-only baselines in terms of average BLEU. This shows that image information as the auxiliary context brings more significant improvement in the massively multilingual translation by nearly +4 BLEU points. The visual features of different languages from ViT encoder is successfully projected into the shared semantic.

\section{Ablation Study}
\paragraph{Performance on Different Backbones.}
In Table~\ref{tab:backbone_ablation}, we compare the results of \ourmethod{} by using the different vision backbones, including ResNet and Transformer~\cite{clip}. In Table~\ref{tab:backbone_ablation}, we observe that \ourmethod{} with the Transformer backbone outperforms the counterpart with CNN network. It shows that our method can unify the two views of visual and language data in the Transformer backbone. Besides, the vision Transformer with smaller patch size (ViT-B/14) gets the better performance but generates longer visual tokens for computation compared to ViT-B/32 and ViT-B/16. Therefore, we recommend the ViT-B/32 for efficiency or ViT-B/16 for performance as the vision encoder backbone.

\paragraph{Effect of Different Modules.}
Table~\ref{module_ablation} summarizes the ablation study of our proposed modules, which shows that each approach has a significant contribution to the final model. Our multilingual model is first trained on the multilingual data, where the model is denoted as {\large{\ding{175}}} in contrast to bilingual model {\large{\ding{176}}}. Given the sentence pair with image, we adopt the visual representations to enhance the translation. The performance of multimodal translation is improved by the alternative training strategy (\drop{}), where the model is randomly trained with visual or language tokens ({\large{\ding{174}}}). Since the source sentences are more important for translation than images, \cvlm{} uses the language tokens as query and visual tokens as (key, value) for cross-attention, which we denoted as {\large{\ding{173}}}. We further introduce MMCL to explicitly narrow the gap among different languages. Putting them all together, we obtain the final model {\large{\ding{172}}} \textbf{\ourmethod{}}, which proves the effectiveness of progressive learning that can gradually improve performance in different aspects.

\section{Analysis}

\paragraph{Distance of Different Languages.}
The image as a universal language is used to narrow the distance among multiple languages, we visualize the sentence representations of the last language encoder layer. We select 500 parallel sentences from the valid set of four languages, including English, German, French, and Czech. Then, we apply t-SNE~\cite{t_SNE} to reduce the 1024-dim representations to 2-dim. It is clear in Figure~\ref{tsne} that text-only MNMT cannot align the 4 languages. By contrast, \ourmethod{} draws the representations across 3 languages much closer.

\paragraph{Low-resource Setting.}
To further analyze the performance of \ourmethod{} given different sizes of downstream parallel data with image context, we randomly extract $P$ percentage of the whole sentence pairs of different languages as the fine-tuned parallel data from the Multi30k dataset. We set $P$ = \{$10\%$, $20\%$, $\dots$, $100\%$\} and compare our method with the text-only MNMT model. Figure~\ref{low_resource} shows the BLEU points of our pre-trained multilingual model and the baseline on four directions, including En$\to$De, En$\to$Fr, De$\to$En, and Fr$\to$En. When the parallel data size is small, the baseline without pre-trained model produces unsatisfactory results. Similarly, in Figure~\ref{low_resource_en2fr}, \ourmethod{} fine-tuned on nearly $90\%$ data defeats the baseline trained on all pairs, exemplifying the effectiveness of our method in low-resource scenarios.

\paragraph{Vision-Language Alignment.}
The function of multimodal multilingual contrastive learning is used to align vision and language, which aims to project vision and language into the same space. In Figure~\ref{cvlm}, we visualize the conditional vision-language alignment (CVLM) between the source sentence of different languages and image patches. For example, Figure~\ref{cvlm} plots the original sentence and Figure~\ref{cvlm_en} shows cross-attention between English sentence ``\textcolor{teal}{A young child is standing alone} on some \textcolor{blue}{jagged rocks}.'' and image patches. Similarly, Figure~\ref{cvlm_de} describes the attention about the German counterpart ``\textcolor{teal}{Ein kleines Kind steht allein} auf einem \textcolor{blue}{zerklüfteten Felsen}.'' We can oberserve that given diffenet sentences with the same meaning tends to pay attention to the similar image regions, such as \textcolor{blue}{jagged rocks} and \textcolor{blue}{zerklüfteten Felsen}.
Figure~\ref{cvlm_en}$\sim$\ref{cvlm_cs} indicate our method effectively force the model to learn the similar vision-language attention pattern and project different languages into the same semantic space using MMCL and CVLM.

\begin{figure}[t]
\centering
    \subfigure[]{
    \includegraphics[width=0.4\columnwidth]{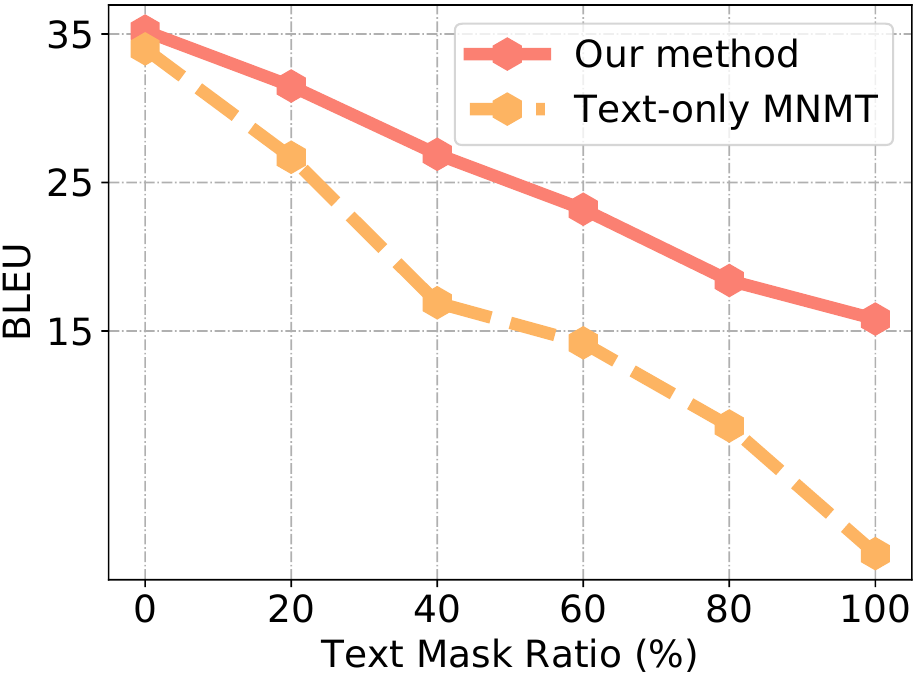}\quad
    \label{mask_ratio_en2cs}
    }
    \subfigure[]{
    \includegraphics[width=0.4\columnwidth]{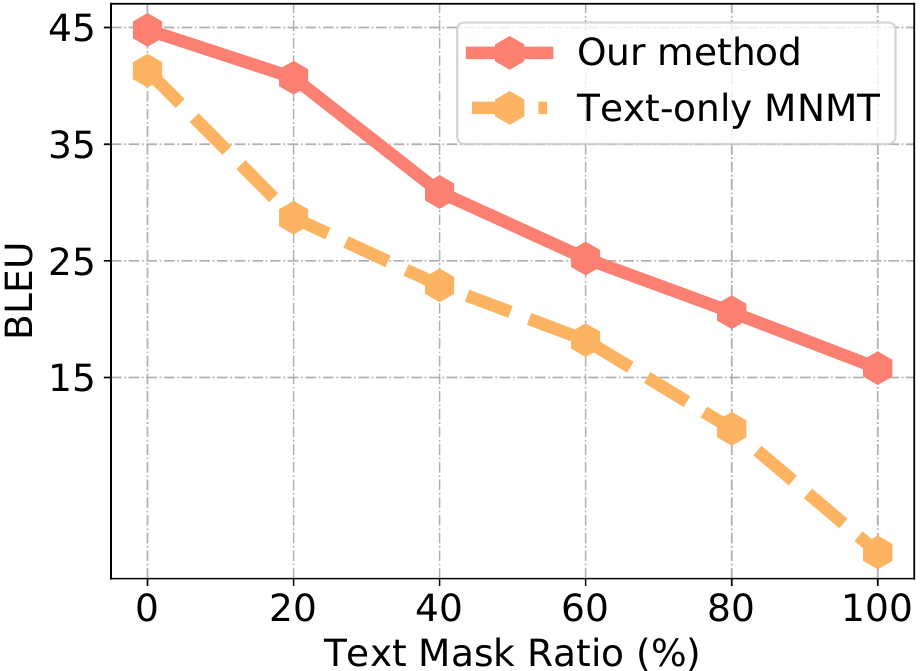}\quad
    \label{mask_ratio_cs2en}
    }
    \caption{Comparison between the text-only MNMT and \ourmethod{} when the source sentence is masked with different ratios.} 
    \vspace{-10pt}
    \label{mask_ratio}
\end{figure}

\paragraph{Sanity Check on Visual Context.}
In Figure~\ref{mask_ratio}, we compare our \ourmethod{} with the text-only multilingual model to emphasize the necessity of visual context by masking source words with different mask ratios \{$0\%$, $20\%$, $40\%$, $60\%$, $80\%$, $100\%$\}. When the source sentence is masked, the visual context provide the supplementary information to help translation correctly. When only receiving the source language, the performance of MNMT is obviously worse than \ourmethod{}, where the visual representations from vision encoder can compensate for the masked words. When the mask ratio is $0\%$, MNMT can not perform translation since the all words are masked while \ourmethod{} outperforms MNMT by a large margin (nearly $15$ BLEU points). Despite that all source words are masked, \ourmethod{} can perform image caption under this extreme scenario due to \drop{}.

\section{Related Work}

\paragraph{Multilingual Multimodal Translation.}
Multilingual Neural Machine Translation (MNMT) aims to support multiple translation directions by sharing parameters. Recent works~\cite{massively_mnmt,massively_mnmt_improved,massively_thousand,microsoft_wmt2021,xmt,um4,hltmt,ganlm,MNMT_low_resource,wang2023mt4crossoie,liu2022cross} scale to the massively multilingual setting to support more languages. Despite these benefits, the multilingual model tends to underperform its bilingual counterparts with worse translation performance~\cite{mnmt_challenges}.
Multimodal machine translation (MMT) refers to the process of translating content that includes both text and images from one language to another, which is a challenging task that aims to enhance source-target translation extra visual context. Researchers propose different attention mechanisms to incorporate language and vision features based on the encoder-decoder architecture~\cite{lium_wmt2018,multimodal_transformer_mmt,graph_mmt,visual_context_mmt,mmt_phrase_vision,on_vision_matters,lvp,m2c} and decoder-only models~\cite{minigpt4}.
Vision-language pre-trained models have the ability to process visual information and understand natural language jointly~\cite{ViLBERT,uniter,vl_bert,layoutxlmv3,clip,videoclip}. These vision-language models~\cite{ViLBERT,lxmert,clip} perform remarkably on various benchmarks and demonstrated to be effective in a range of tasks, including image and video captioning~\cite{clip4video}, visual question answering~\cite{vlmo,ofa,beit3}, and multimodal machine translation~\cite{probing_mmt}. 

\paragraph{Large Language Model.}
Large language models (LLM)~\cite{liu20242,wang2023rolellm} have emerged as a significant milestone in the field of natural language processing, such as GPT~\cite{gpt4}, OPT~\cite{opt}, Llama~\cite{llama,llama2}, BLOOM~\cite{bloom}. These models demonstrated remarkable proficiency in understanding and generating human language, offering the potential for a wide range of applications in fields such as natural language understanding, text generation, and conversational AI. Instruction tuning
(IT) is proposed to align the LLM to follow instructions response~\cite{crossmodal_prompt,sft_survey,flan_moe,self_instruct} and bridge the gap between the next-word prediction objective and the downstream tasks.

\section{Conclusion}

In this work, we introduce \ourmethod{}, a state-of-the-art multilingual multimodal machine translation model, which supports multiple translation directions of 102 languages guided by image context.
To narrow the gap among different languages, the image is operated as the central language by contrastive learning (MMCL) trained on the multilingual text-image pairs.
Then, we incorporate the visual context into the language representations as the conditional vision-language memory (CVLM) for multilingual generation.
Extensive experiments prove the effectiveness of \ourmethod{} on the Multi30k and the extended large-scale dataset \dataset{} of 102 languages. The importance of visual signals in multilingual training has been further verified by a series of probing experiments.

\section*{Acknowledgements}

This work was supported in part by the National Natural Science Foundation of China (Grant Nos. U1636211, U2333205, 61672081, 62302025, 62276017), a fund project: State Grid Co., Ltd. Technology R\&D Project (ProjectName: Research on Key Technologies of Data Scenario-based Security Governance and Emergency Blocking in Power Monitoring System, Proiect No.: 5108-202303439A-3-2-ZN), the 2022 CCF-NSFOCUS Kun-Peng Scientific Research Fund and the Opening Project of Shanghai Trusted Industrial Control Platform and the State Key Laboratory of Complex \& Critical Software Environment (Grant No. SKLSDE-2021ZX-18).




\nocite{*}
\section{Bibliographical References}\label{sec:reference}

\bibliographystyle{lrec-coling2024-natbib}
\bibliography{m3p}


\end{document}